%% file: main_arxiv.tex
\documentclass[runningheads]{llncs}
\usepackage{graphicx}
\usepackage{tikz}
\usepackage{comment}
\usepackage{amsmath,amssymb} 
\usepackage{color}
\usepackage{subfiles}
\usepackage{multirow}
\usepackage{amsmath}
\usepackage{booktabs}  
\usepackage{xcolor,pifont}
\usepackage{fontawesome}
\usepackage{subcaption}
\definecolor{mygreen}{RGB}{0, 153, 0}

\usepackage[accsupp]{axessibility}  
\newcommand{\Paragraph}[1]{\vspace{0mm} \noindent \textbf{#1} \hspace{0mm}}
\usepackage{amsmath}

\DeclareMathOperator*{\argmin}{arg\,min}

\newcommand{\greencheck}{\textcolor{green}{\ding{51}}} 
\newcommand{\redX}{\textcolor{red}{\ding{55}}}

\usepackage{cite}
\usepackage[pagebackref,breaklinks,colorlinks]{hyperref}

\begin{document}
\pagestyle{headings}
\mainmatter
\def\ECCVSubNumber{2888}  

\title{2D GANs Meet Unsupervised Single-view 3D Reconstruction}

\titlerunning{2D GANs Meet Unsupervised Single-view 3D Reconstruction}
%
\author{Feng Liu, Xiaoming Liu}
\authorrunning{F. Liu et al.}
\institute{Michigan State University, Computer Science \& Engineering\\
\email{\{liufeng6,liuxm\}@msu.edu}}

\maketitle

\subfile{sec_0_abstract.tex}


\subfile{sec_1_intro.tex}

\subfile{sec_2_priors.tex}

%

\subfile{sec_3_method.tex}

\subfile{sec_4_exp.tex}

\subfile{sec_5_conclusion.tex}

\clearpage
%
%
\bibliographystyle{splncs04}
\bibliography{egbib}
\end{document}

%% file: sec_0_abstract.tex
\begin{abstract}
 
 Recent research has shown that controllable image generation based on pre-trained GANs can benefit a wide range of computer vision tasks.
 However, less attention has been devoted to 3D vision tasks. In light of this, we propose a novel image-conditioned neural implicit field, which can leverage 2D supervisions from GAN-generated multi-view images and perform the single-view reconstruction of generic objects. Firstly, a novel offline StyleGAN-based generator is presented to generate plausible pseudo images with full control over the viewpoint. Then, we propose to utilize a neural implicit function, along with a differentiable renderer to learn 3D geometry from pseudo images with object masks and rough pose initializations. To further detect the unreliable supervisions, we introduce a novel uncertainty module to predict uncertainty maps, which remedy the negative effect of uncertain regions in pseudo images, leading to a better reconstruction performance. The effectiveness of our approach is demonstrated through superior single-view 3D reconstruction results of generic objects. Code is available at \url{http://cvlab.cse.msu.edu/project-gansvr.html}.

\keywords{2D GANs, Multi-view Pseudo Images, Unsupervised, Single-view 3D Reconstruction, Generic objects, Uncertainty }
\end{abstract}

%% file: sec_1_intro.tex
\section{Introduction}\label{sec:intro}

Realistic image synthesis is an important research area of computer vision. There has been remarkable progress in this field with the advent of $2$D Generative Adversarial Networks (GANs)~\cite{goodfellow2014generative}, such as StyleGAN~\cite{karras2019style} and its variations~\cite{karras2020analyzing,Karras2020ada,karras2021alias}, which can generate high-fidelity images of diverse object categories with a wide variety of attributes (\emph{e.g.,} pose, identity).  
Such superior capabilities of modeling the semantic image 
manifold enable the generated photorealistic images to be leveraged for many vision tasks, such as image editing~\cite{karras2017progressive,suzuki2018spatially}, domain translation~\cite{zhu2017unpaired}, face recognition~\cite{controllable-and-guided-face-synthesis-for-unconstrained-face-recognition}, and video generation~\cite{tulyakov2018mocogan}.
However, it remains much less explored in $3$D vision tasks, \emph{e.g.,} $3$D reconstruction~\cite{liu2021fully,liu2021voxel}.

The $2$D GAN manifolds appear to learn $3$D geometrical properties implicitly, where recent GAN interpretation methods~\cite{shen2021closed,harkonen2020ganspace} have shown that manipulating the latent code of the pre-trained GAN models can produce images of the same object under different viewpoints. 
Our work aims to answer the following question.
\emph{Using the GAN-generated multi-view images, can we learn a category-specific multi-view stereo system without $3$D supervision that can reconstruct $3$D shapes from a single image?}

\begin{figure}[t]
\centering
\includegraphics[trim=0 0 0 0,clip, width=0.90\linewidth]{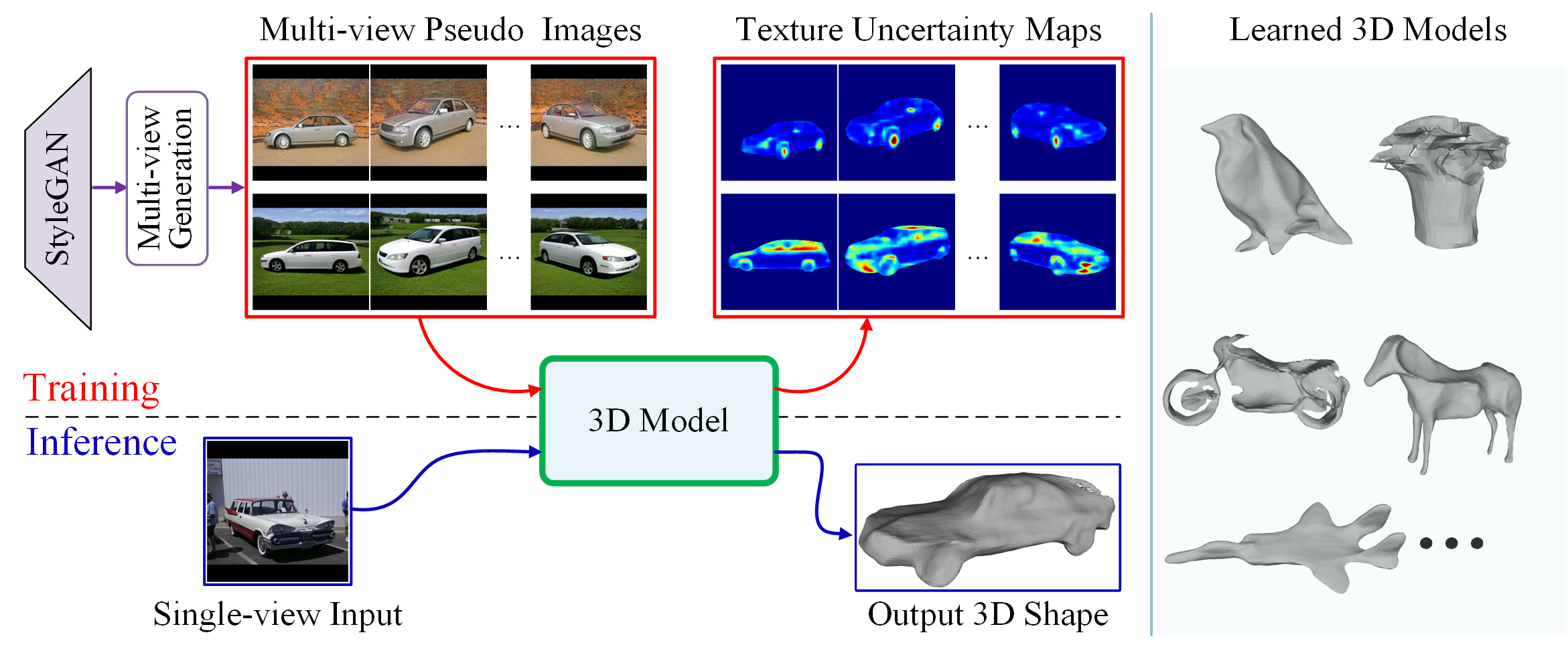}
\caption{ Our approach leverages StyleGAN-generated multi-view pseudo images to learn a $3$D model without $3$D supervision, which can perform single-view $3$D reconstruction for a variety of generic objects, \emph{e.g.}, airplanes, birds, cars, horses, motorbikes, potted plants, \emph{etc}. In addition, our framework produces uncertainty maps, indicating the unreliable local areas in the pseudo images.}
\label{fig:teaser}
\end{figure}

Early attempts~\cite{wu2020unsupervised,pan20202d,shi2021lifting} are made to mine $3$D geometric cues from the pre-trained $2$D GAN models in an unsupervised manner. 
However, without modeling objects in the $3$D space, these methods only recover $2.5$D representations (depth or normals). 
Recently, StyleGANRender~\cite{zhang2020image} integrates StyleGAN to generate multi-view images, which may be used to train an inverse graphics network for $3$D reconstruction.
However, the method focuses more on performing independent manipulation of $3$D properties in GAN's latent space by fine-tuning the GAN models. In addition, the unreliable texture existing in GAN-generated multi-view images is a common issue, and has not been investigated.

It remains a challenge to leverage GAN-generated multi-view supervision for single-view $3$D reconstruction.
First of all, the pre-trained $2$D GAN models lack explicit and precise camera pose to control over generated images, which is a necessity for classic multi-view stereo. Second, the GAN-generated multi-view images often suffer from {\it local} distortion and low perceptual quality, which severely breaks the {\it consistency} of either object shape or texture across views, and thereby ruins the {\it cornerstone} of multi-view stereo.

To address these challenges, we propose a novel framework to leverage GAN-generated multi-view images (termed `\emph{pseudo images}') in learning generic object shape models, for the purpose of $3$D reconstruction from a single image (Fig.~\ref{fig:teaser}). 
To first generate multi-view imagery by a pre-trained GAN, {\it e.g.}, StyleGAN, we carefully study the latent space of StyleGAN and devise a simple but effective technique, which generates plausible images with an azimuth range of $0{-}360^{\circ}$. 
Consequently, during training,  given a realistic image generated by StyleGAN, we can produce a set of pseudo images of the same object under different viewpoints. We then introduce a neural implicit network to simultaneously learn the unknown geometry, texture, and camera parameters for the objective of reconstructing the pseudo images, by incorporating a differentiable renderer. 
%

A key component is that we introduce a learning framework that enables the neural implicit network to be conditioned on a single image. 
Specifically, we adopt an image encoder as a hypernetwork to predict the network parameters of the implicit function.  
This image conditioning allows the framework to be trained on multi-view images, where it learns object geometry priors within the category to perform single-view reconstruction. Moreover, to address the unreliable texture supervision issue in pseudo images, we devise an uncertainty prediction module, together with an uncertainty-aware photometric loss to estimate uncertainty maps, which can effectively filter out the unreliable supervision signals\textbf{/}inconsistencies within\textbf{/}across multi-view pseudo images, leading to a more precise reconstruction. 
Comprehensive experiments show the superiority of our method over existing methods in unsupervised single-view $3$D reconstruction.

In summary, the contributions of this work include:

$\diamond$ We propose a novel image-conditioned neural implicit network, which can exploit $2$D supervision from GAN-generated multi-view pseudo images and performs single-view $3$D reconstruction of generic objects. 

$\diamond$ We introduce a multi-view image generation mechanism based on the pre-trained StyleGAN models, which can produce plausible images with full control over viewpoints.     
    
$\diamond$ We propose an uncertainty prediction module to ignore unreliable texture supervision in pseudo images, enabling a reliable self-supervised learning.
    
$\diamond$ Our method shows superior single-view $3$D reconstruction for rigid and non-rigid generic objects in the wild.

%% file: sec_2_priors.tex
\section{Prior Work}\label{sec:prior}

\Paragraph{Application of Pre-trained $2$D GANs}
While research on GANs is rapidly growing, 
our review mainly focuses on the pre-trained unconditional $2$D GAN models.
The capability to produce high-quality images makes $2$D GANs applicable to many vision tasks, {\it e.g.}, image restoration~\cite{wang2021towards,yang2021gan}, image editing (inpainting, super-resolution, semantic manipulation)~\cite{gu2020image,pan2021exploiting}, segmentation~\cite{zhang2021datasetgan}, and DeepFake attack and defense~\cite{dang2020detection,rossler2019faceforensics++,dolhansky2020deepfake,asnani2021reverse}.
Further, the pre-trained $2$D GAN models have been applied to data augmentation to reduce overfitting and bias in deep models~\cite{su2020pre,rojtberg2020style}.
To expand to $3$D vision applications, prior works~\cite{wu2016learning,lunz2020inverse,szabo2019unsupervised,wu2020unsupervised,zhu2018visual} adopt GANs to learn $3$D shapes from images but rely on either $3$D supervision or a $3$D generator, which suffers from heavy memory consumption or extra training difficulties. 
 Recently, 
 LiftedGAN~\cite{shi2021lifting} lifts a pre-trained StyleGAN and distill it into a $3$D aware generator, producing depth maps as a by-product. 
Similarly, GAN2Shape~\cite{pan20202d} produces an unsupervised decomposition by using a GAN model as supervision.
However, those methods require inefficient online image generation during training and infer $2.5$D representations only. 
StyleGANRender~\cite{zhang2020image} 
exploits StyleGAN as a multi-view generator to learn an \emph{mesh-based} inverse graphics network 
to turn the StyleGAN into a controllable render.
However, they require a \emph{fine-tuning} step for the entire StyleGAN model, which is not desirable in this work.
Moreover, they do not tackle the unreliability in pseudo multi-view images.
In contrast, our method focuses on leveraging \emph{pre-trained} $2$D GANs for single-view $3$D reconstruction. Despite both methods utilizing GAN-generated pseudo images for $3$D modeling, we step forward in more plausible multi-view generation, robust shape and texture representation, and uncertainty-aware photometric supervision mechanism.

\begin{table}[t!]
\newcommand{\tabincell}[2]{\begin{tabular}{@{}#1@{}}#2\end{tabular}}
\centering
\caption{{ Comparison of \textbf{\emph{unsupervised}} shape learning methods. [Keys: 
Cam. = camera poses per training sample,  Requ.~or Cons.~= requirement or constraint, Real data= whether can train on real-world images, \textcolor{red}{\faTimesCircleO}= camera poses for a set of reference images]}}
\resizebox{1\linewidth}{!}{
\begin{tabular}{l |c| c | c | c | c  }
\hline
Method &  Output Representation &  Required Template    & Required Cam.    & Additional Requ.~or Cons.  & Real data\\
\hline\hline
LiftedGAN~\cite{shi2021lifting} & $2.5$D, depth & \redX & \redX &  GAN models, pre-trained &  \greencheck\\
GAN2Shape~\cite{pan20202d} & $2.5$D, depth & \redX & \redX & GAN models, pre-trained & \greencheck  \\
StyleGANRender~\cite{zhang2020image} & $3$D, mesh & \redX & \redX & GAN models, fine-tuning & \greencheck \\

DVR~\cite{niemeyer2020differentiable}   & $3$D, implicit & \redX & \greencheck & multi-view & \redX  \\
DIST~\cite{liu2020dist}    & $3$D, implicit & \redX & \greencheck & multi-view & \redX  \\
SDFDiff~\cite{jiang2020sdfdiff} & $3$D, implicit & \redX & \greencheck & multi-view &  \redX \\

CSDM~\cite{tulsiani2016learning}    & $3$D, mesh & \greencheck & \greencheck & $2$D semantic &  \greencheck \\
 CMR~\cite{kanazawa2018learning}    & $3$D, mesh & \greencheck & \redX & $2$D semantic &  \greencheck \\
U-CMR~\cite{goel2020shape}   & $3$D, mesh & \greencheck & \redX & viewpoint distribution   &  \greencheck \\
UMR~\cite{li2020self}   & $3$D, mesh & \redX & \redX & $3$D semantic    &  \greencheck \\
CSM~\cite{kulkarni2019canonical}   & $3$D, mesh & \greencheck & \redX & - &  \greencheck \\
A-CSM~\cite{kulkarni2020articulation}   & $3$D, mesh & \greencheck  & \redX & - &  \greencheck \\
DRC~\cite{tulsiani2017multi}   & $3$D, voxel & \redX  & \greencheck & multi-view &  \greencheck \\

 SRN~\cite{sitzmann2019scene}   & $3$D, implicit & \redX & \greencheck & multi-view & \redX  \\
 NeRF~\cite{mildenhall2020nerf}  & $3$D, implicit & \redX & \greencheck & multi-view & \greencheck  \\
 SDF-SRN~\cite{lin2020sdf}  & $3$D, implicit & \redX & \greencheck & - & \greencheck  \\
 ShSMesh~\cite{ye2021shelf}  & $3$D, volumetric & \redX & \redX & - & \greencheck  \\
\hline

Proposed & $3$D, implicit & \redX & \textcolor{red}{\faTimesCircleO} & GAN models, pre-trained &  \greencheck \\
\hline
\end{tabular}
}
\label{tab:3D_modeling_review}
\end{table}

\Paragraph{$3$D-aware Generative Models}
Understanding the latent representation of GANs has 
resulted in a body of works disentangling various factors of generated objects
in a $3$D-controllable manner, \emph{e.g.}, viewpoint.
These approaches can be classified into two groups. 
One adds additional modules or losses in training to explicitly disentangle $3$D factors~\cite{most-gan-3d-morphable-stylegan-for-disentangled-face-image-manipulation}. 
For example, HoloGAN~\cite{nguyen2019hologan} controls the object pose by rigid-body transformations via a $3$D feature module. 
StyleFlow~\cite{abdal2021styleflow} learns non-linear paths in the latent space by normalizing flows conditioned on the attribute.  
Recently there has been a trend combining of the neural radiance fields (NeRF)~\cite{schwarz2020graf,niemeyer2021campari,niemeyer2021giraffe,gu2021stylenerf,chan2021pi} with GANs to devise $3$D-aware generators. 
Another line of works, such as InterFaceGAN~\cite{shen2020interfacegan}, SeFa~\cite{shen2021closed}, GANSpace~\cite{harkonen2020ganspace}, discover the latent semantic directions of a pre-trained GAN model that can manipulate object rotation unaware of its underlying $3$D model. 
%
It is preferable to exploit the knowledge contained in a pre-trained GAN image manifold for the goal of recovering $3$D object shapes without retraining the GAN models.

\Paragraph{Shape Learning without $3$D Supervision}
While $3$D reconstruction, especially for faces~\cite{joint-face-alignment-and-3d-face-reconstruction-with-application-to-face-recognition, on-learning-3d-face-morphable-model-from-in-the-wild-images,bai2020deep}, is a long-standing topic, we focus our review on shape learning from real-world images of generic objects without $3$D supervision.
Recent neural networks tackle this ill-posed problem via a differentiable renderer along with a choice of $3$D shape representation~\cite{liu2019soft,jiang2020sdfdiff,liu2020dist,niemeyer2020differentiable,tulsiani2017multi,sitzmann2019scene}. 
However, in these works, 
multiple views of the same object with known cameras are required, which limits their learning from real-world images. 
Another branch of works show promising reconstruction from real-world images~\cite{goel2020shape,li2020self,kulkarni2019canonical,kulkarni2020articulation,wu2021shape}. 
SDF-SRN~\cite{lin2020sdf} mines more supervision from $2$D silhouette for superior reconstruction, yet still requires camera pose.  
ShSMesh~\cite{ye2021shelf} further discards the need for former constraints, but their reconstructions are of lower quality. 
NeRF~\cite{mildenhall2020nerf} and its variations
are scene- or object-specific models, which limit their applications for $3$D reconstruction from unseen objects or scenes. 
In contrast, our models are category-specific, and can perform single-view $3$D reconstruction for novel instances.
Tab.~\ref{tab:3D_modeling_review} summarizes the differences between our method and prior work.

%% file: sec_3_method.tex
\begin{figure}[t]
\centering
\includegraphics[trim=0 6 0 0,clip, width=0.95\linewidth]{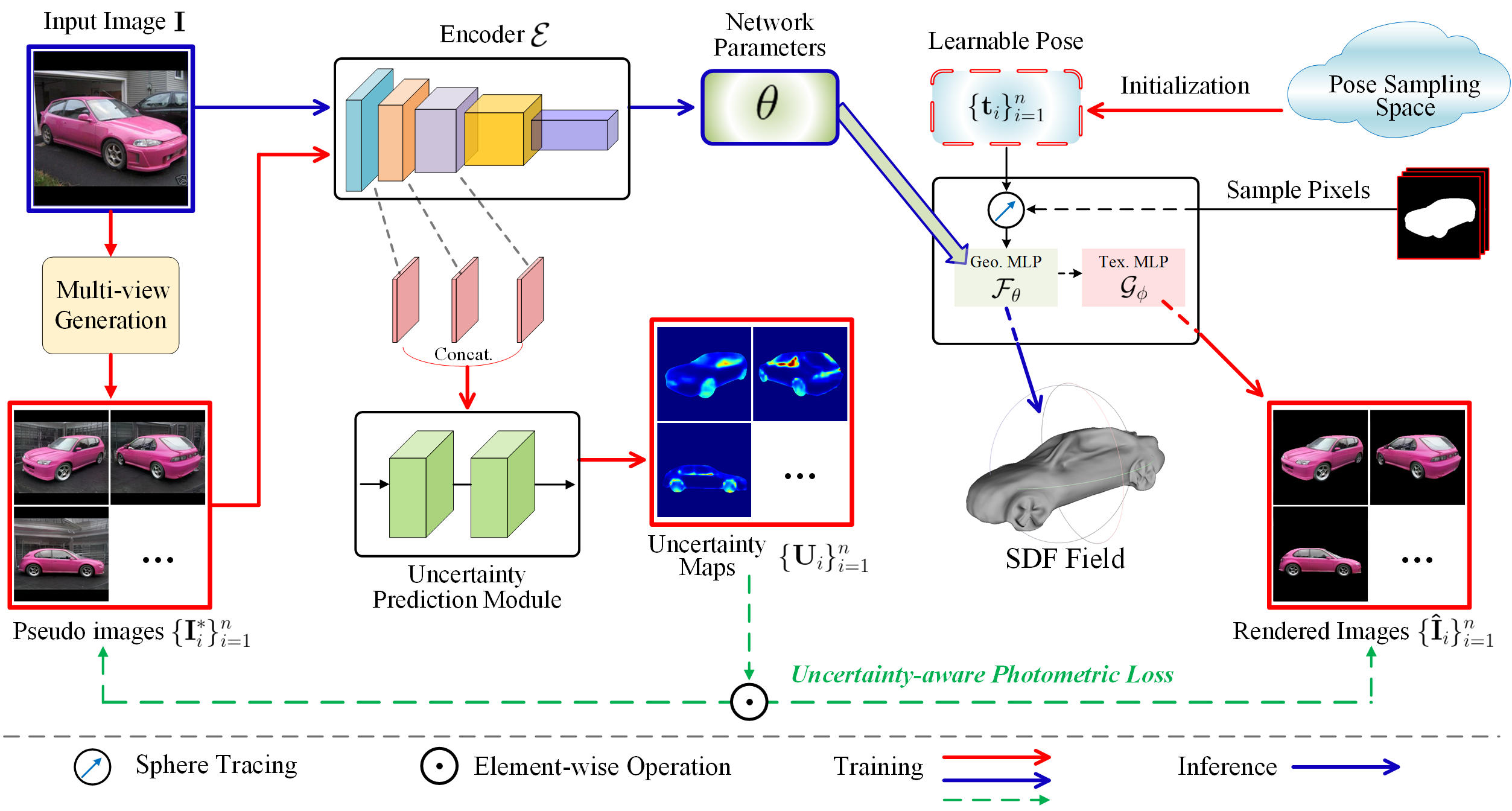}
\caption{\textbf{Overview.} The proposed framework is composed of two key modules: an offline StyleGAN-based multi-view generation and an image-conditioned neural implicit network. During training, the neural implicit network learns unknown geometry, texture, and camera poses for the objective of approximating the multi-view pseudo images. At inference time, the learned neural implicit function performs $3$D reconstruction for the object from a single image.}
\label{fig:flowchart}
\vspace{-2mm}
\end{figure}

\section{Proposed Method}
We start with an overview of the proposed framework (Fig.~\ref{fig:flowchart}) and then present the individual modules in detail.
We first introduce an offline and effective multi-view generator based on the pre-trained StyleGAN models, which produce plausible multi-view images with full control over viewpoints. 
Then, we detail the proposed image-conditioned neural implicit field learning framework, including a neural implicit network, differentiable rendering procedure, and an uncertainty prediction module. These three modules work jointly for the objective of exploiting the pseudo images to learn generic object shape priors and perform $3$D reconstruction from a single input image.

\subsection{StyleGAN based Multi-view Generation}~\label{sec:multi-view}
\begin{figure}
     \centering
     \begin{subfigure}[b]{0.48\textwidth}
         \centering
         \includegraphics[width=\textwidth]{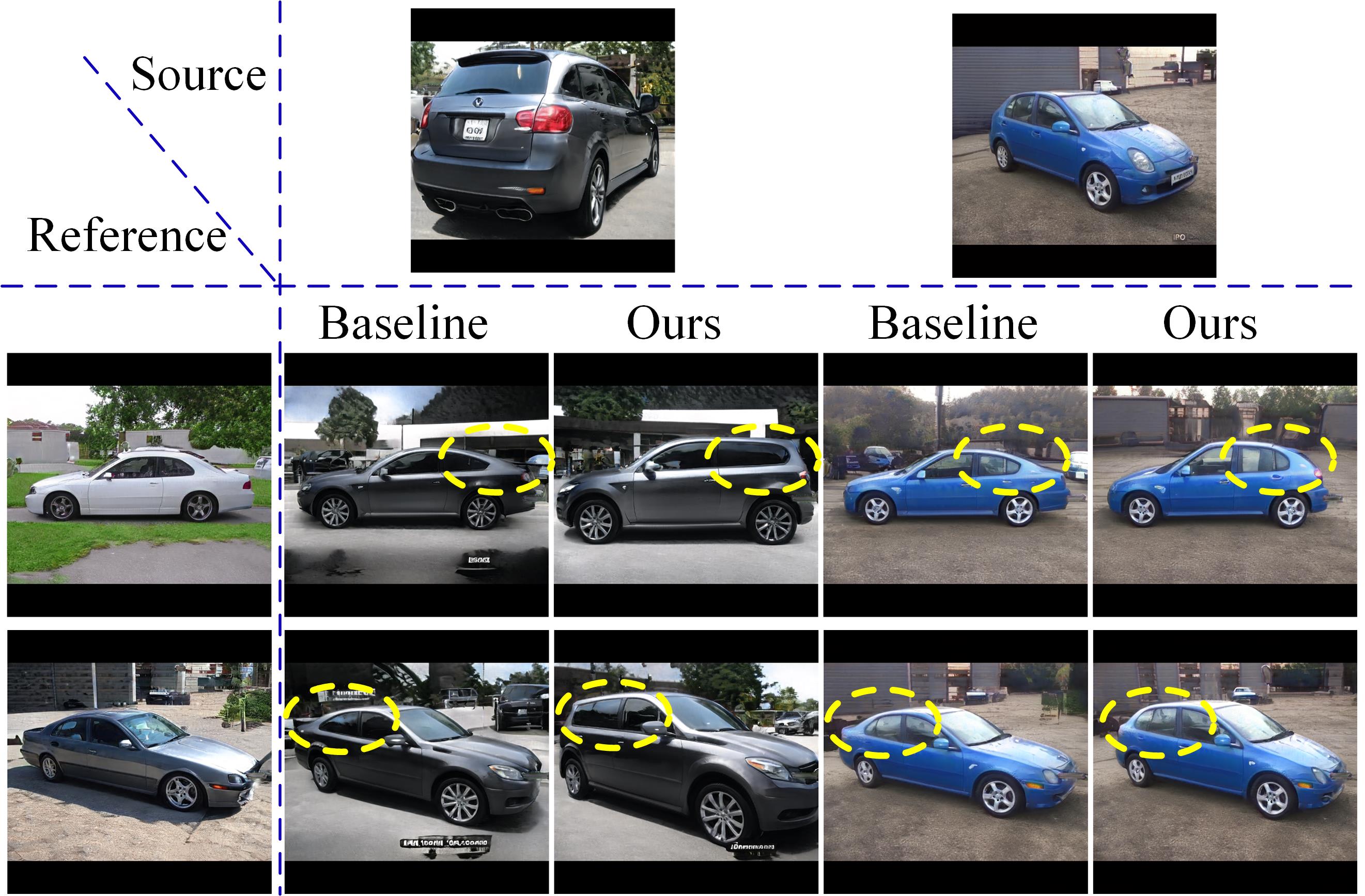}
         \caption{}
         \label{fig:multi_view}
     \end{subfigure}
     \hspace{2mm}
     \begin{subfigure}[b]{0.48\textwidth}
         \centering
         \includegraphics[width=\textwidth]{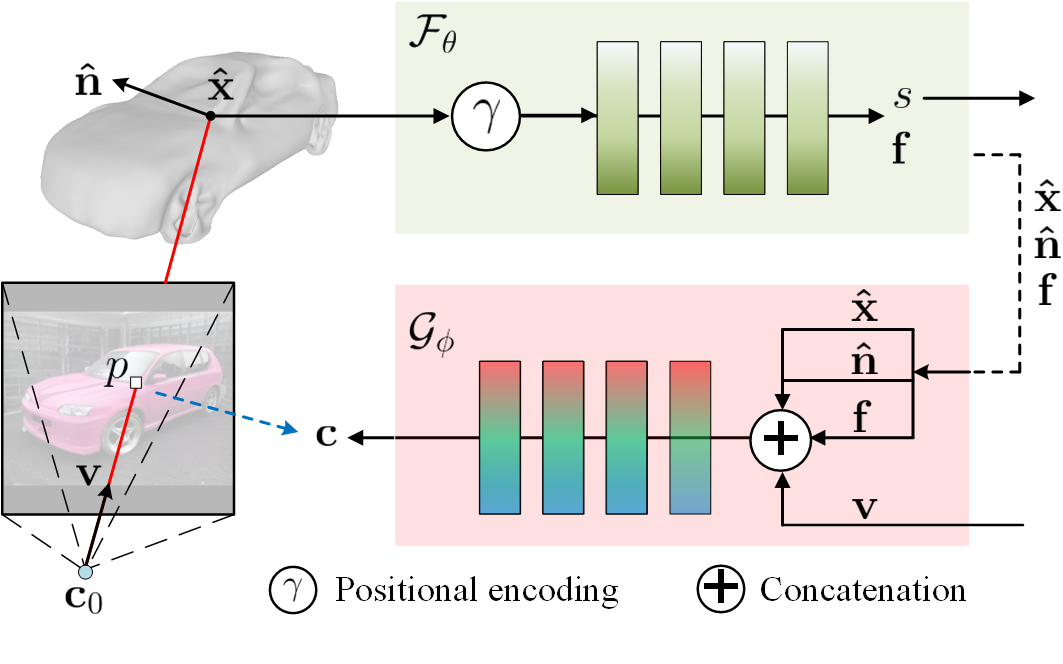}
         \caption{}
         \label{fig:implicit}
     \end{subfigure}
        \caption{ (a) Multi-view generator comparisons of \textbf{\emph{Baseline}}~\cite{zhang2020image} and our approach (\textbf{\emph{Ours}}). As can be observed, besides the viewpoint, the baseline perceives shape cues from the references (\textcolor{yellow}{yellow} circle, best view in zoom in). While our generated images show more consistency in object shape across views, which benefits shape learning. Please refer to \textbf{Supp} for more categories. (b) The architecture of our neural implicit fields and differentiable renderer. }
\end{figure}

We briefly review the embedding space of the StyleGAN~\cite{karras2019style,karras2020analyzing}. 
Typically, a generator $G(\cdot)$ samples a latent code $\mathbf{z}$ from a pre-defined distribution $\mathcal{Z}$ such as the normal distribution, and produces an output image $\mathbf{I}$. 
The code $\mathbf{z}$ is first mapped to an intermediate latent space $\mathcal{W}$ via a Multi-layer Perceptron (MLP), and then $\mathcal{W}$ is transformed to $\mathcal{W}^{+}$ space by $16$ learned affine transformations.
The generator $G(\cdot)$ projects $\mathbf{W}$ to the final image: $G(\mathbf{W})\!=\!\mathbf{I}$.
Such latent codes have been shown to learn various disentangled semantics~\cite{shen2021closed,harkonen2020ganspace}.
For instance, StyleGANRender~\cite{zhang2020image} finds that the latent codes $\mathbf{W}_{v}\!:=\!(\mathbf{w}_1,\mathbf{w}_2,\mathbf{w}_3,\mathbf{w}_4)\!\in \!\mathbb{R}^{4\times 512}$ in the first $4$ layers control camera viewpoints. 
That is, given a source and reference generated image pair $(\mathbf{I}^S,\mathbf{I}^R)$ with their latent codes $(\mathbf{W}^{S},\mathbf{W}^{R})$, we can generate an image of the source object with the reference viewpoint by swapping $(\mathbf{W}_v^S, \mathbf{W}_v^R)$ and keeping the rest dimensions of $\mathbf{W}^{S}$. We denote this multi-view generation strategy as \textbf{\emph{Baseline}}.
However, while $\mathbf{W}_{v}$ indeed alters the object viewpoint, it still perceives the shape of the reference object, as shown in Fig.~\ref{fig:multi_view}.

It is difficult to develop a multi-view stereo system from noisy multi-view images with inconsistent shapes. 
To tackle this issue, inspired by SeFa~\cite{shen2021closed}, we propose a novel offline multi-view generation mechanism that generates plausible images with enhanced cross-view consistency.
SeFa suggests that the weight parameters in early affine transformations contain essential knowledge of image variations. 
One can obtain interpretable directions in the latent space by computing eigenvectors of their weight matrices and selecting eigenvectors with the largest eigenvalues. 
With this observation, we propose to filter the viewpoint-irrelevant features in $\mathbf{W}_{v}^{R}$ guided by the eigenvectors of the $k$ largest eigenvalues, which are computed from the weight parameters in the first $4$ transformations.
It can be formulated as:
\begin{equation}
    \argmin_{\alpha}||\hat{\mathbf{W}}_{v} - \mathbf{W}_{v}^{R}||^2, \quad \hat{\mathbf{W}}_{v} = \alpha\mathbf{V}+\mathbf{W}_{v}^{S}, 
    \label{eqn:w_latent_code}
\end{equation}
where $\hat{\mathbf{W}}_{v}$ is the enhanced viewpoint latent code.
$\mathbf{\alpha}$ is a $k$-dim viewpoint coefficient. $\mathbf{V}\in \mathbb{R}^{k\times 512}$ denotes the eigenvectors of $\mathbf{A}^T\mathbf{A}$ associated with the $k$ largest eigenvalues. 
$\mathbf{A}\in \mathbb{R}^{m\times 512}$ are the weights of transformations. $\alpha\mathbf{V}$ is duplicated into four rows in Eqn.~\ref{eqn:w_latent_code} and $\alpha$ can be solved by gradient descent. Finally, we generate a new image by combining $\hat{\mathbf{W}}_{v}$ and remaining the dimensions of $\mathbf{W}^{S}$.

\Paragraph{Training Pseudo Images} Given a pre-trained StyleGAN model, we first synthesize training images and filter out images that have more than one instance or an unrealistic instance, resulting in $N$ training samples $\{\mathbf{I}^{j}\}_{j=1}^{N}$. 
Then, we manually select $n$ reference view samples, which roughly cover the common object viewpoints ranging from $0-360^{\circ}$ in azimuth. 
Finally, for each training sample $\mathbf{I}^{j}$, we produce $n$ multi-view images of the same object with fixed camera poses as the pseudo images $\{\mathbf{I}^{*j}_{i}\}_{i=1}^{n} \in \mathbb{R}^{W\times H\times 3}$. We further apply the work~\cite{qin2020u2} to obtain corresponding instance segmentation $\{\mathbf{M}^{j}_{i}\}_{i=1}^{n}$ of pseudo images.

\Paragraph{Pose Sampling Space} 
We assume a pinhole camera model $\mathcal{C}=(\mathbf{t},\mathbf{K})$, where $\mathbf{K}\in\mathbb{R}^{3\times3}$ is the intrinsic parameter. 
We represent the camera pose/extrinsic parameters $\mathbf{t}=( \mathbf{c}_{0},\mathbf{q})$ based on its $3$D position $ \mathbf{c}_{0}\in\mathbb{R}^3$ and its rotation from a canonical view. $\mathbf{q}\in\mathbb{R}^4$ is the quaternion vector representing the camera rotation. We assume the observed object is approximately inside the unit sphere. We further assume a known intrinsic and the principal point at the image center, as commonly assumed in stereo systems~\cite{kar2017learning}. 
For the pose initialization,  we manually annotate the camera poses $\{\mathbf{\tilde{t}}_{i}\}_{i=1}^n$ for the $n$ reference images, which initializes the pseudo images' poses, $\mathbf{t}_{i}^j=\mathbf{\tilde{t}}_{i}, j\!\in\![1,N]$. 
\emph{Since we only need to annotate $n$ reference samples per object category, it is far more practical than prior works that require camera pose label per training sample} (see Tab.~\ref{tab:3D_modeling_review}).
%

\subsection{Image-conditioned Neural Implicit Field}
\Paragraph{$3$D Geometry Representation} 
As illustrated in Fig.~\ref{fig:implicit}, the object geometry to be reconstructed is represented by the function~\cite{yariv2020multiview}: $\mathcal{F}: \gamma(\mathbf{x}) \rightarrow (s, \mathbf{f}) $ that maps a point $\mathbf{x} \in\mathbb{R}^3$ to its signed distance value $s$ to the object surface and a local geometry feature $\mathbf{f}\in \mathbb{R}^{d_{f}}$. $\gamma (\cdot)$ denotes a positional encoding operator on $\mathbf{x}$ with $6$ exponentially increasing frequencies introduced in NeRF~\cite{mildenhall2020nerf}. The surface $\mathcal{S}$ is represented as the zero level set of MLP $\mathcal{F}$ with learnable parameters $\theta$:
\begin{equation}
    \mathcal{S}=\{\mathbf{x}\in \mathbb{R}^3|\mathcal{F}^{(s)}_{\theta}(\gamma (\mathbf{x}))=0\}.
\end{equation}
.

\Paragraph{Neural Renderer}
Given a pixel $p$ of a masked input image, we march a ray $\mathbf{r}=\{\mathbf{c}_0+t\mathbf{v}|t\geqslant 0\}$, where $\mathbf{c}_0$ is the camera position and $\mathbf{v}$ the viewing direction. $\mathbf{\hat{x}}$ denotes the first intersection between the ray $\mathbf{r}$ and the surface, which can be efficiently detected via the sphere tracing~\cite{hart1996sphere} and implemented in a differentiable manner. 
As in Fig.~\ref{fig:implicit}, the rendered color of the pixel $p$ is encoded as~\cite{yariv2020multiview}
\begin{equation}
    \mathcal{G}: (\mathbf{\hat{x}},\mathbf{\hat{n}},\mathbf{f},\mathbf{v}) \rightarrow \mathbf{c},
    \label{eqn:g}
\end{equation}
which is a function of the surface properties at $\mathbf{\hat{x}}_p$, including the surface point $\mathbf{\hat{x}}_p$, the surface normal $\mathbf{\hat{n}}_p\in\mathbb{R}^3$, the local geometry feature $\mathbf{f}_p$, and a viewing direction $\mathbf{v}_p\in\mathbb{R}^3$. 
Similarly, the function $\mathcal{G}$ is implemented as an MLP with learnable parameters $\phi$. 
The surface normal $\mathbf{\hat{n}}_p$ can be computed by the spatial derivative $\frac{\delta \mathcal{F}^{(s)}}{\delta \mathbf{x}_p}$ via back-propagation through the network $ \mathcal{F}$. Incorporating the surface normal and view direction enables $\mathcal{G}$ to represent the light reflected from a surface point $\mathbf{x}$ across different viewpoints~\cite{mildenhall2020nerf,yariv2020multiview}.
Also, according to~\cite{yariv2020multiview}, introducing the local geometry feature vector $\mathbf{f}$ allows the renderer to handle more complex appearances, which might appear in GAN-generated images.
The neural implicit field can be learned through back-propagation without any $3$D supervision by comparing the rendered images with the multi-view pseudo images. Minimizing this error encourages multi-view photo consistency because only when the point is on the actual surface, the MLPs $(\mathcal{F}, \mathcal{G})$ will predict accurate color with fewer multi-view variations. 


\Paragraph{Image Encoder}
We propose to utilize architecture to condition on a single image, such that the learned neural implicit field can generalize to a new object instance without re-training. 
Specifically, as shown in Fig.~\ref{fig:flowchart}, we use an image encoder $\mathcal{E}$ as a hyper network~\cite{ha2016hypernetworks,lin2020sdf} to predict $\theta$ (the parameters of $\mathcal{F}$), written as $\theta=\mathcal{E}_{\Phi}(\mathbf{I})$, where $\Phi$ is the neural network weights.

\subsection{Model Learning}
Given $N$ sets of masked pseudo images $\{\mathbf{I}^{j}, \{\mathbf{I}^{*j}_{i}\}_{i=1}^{n}\}_{j=1}^{N}$, along with their initial camera poses $\{\{\mathbf{t}^{j}_{i}\}_{i=1}^{n}\}_{j=1}^{N}$, we optimize the encoder parameters $\Phi$, texture MLP parameters $\phi$ and camera poses $\{\{\mathbf{t}^{j}_{i}\}_{i=1}^{n}\}_{j=1}^{N}$ by minimizing the  loss:
\begin{equation}
    \mathcal{L} = \sum_{j=1}^{N} (\mathcal{L}_{RGB}+\lambda_{mask}\mathcal{L}_{Mask}+\lambda_{eik}\mathcal{L}_{Eik}),
\end{equation}
where $\mathcal{L}_{RGB}$ is photometric loss, $\mathcal{L}_{Mask}$ is silhouette loss, $\mathcal{L}_{Eik}$ is Eikonal regularization, and $\lambda_*$ are loss weights.

\Paragraph{Photometric Loss}
Let $\mathbf{I}^*_p$, $\mathbf{M}_p\!\in\!\{0,1\}$ be the RGB and silhouette values of pixel $p$ in an image sample $\mathbf{I}^{*}_{i}$ taken at the view direction $\mathbf{v}_p$ associated with camera $\mathcal{C}_{i}$. $p\in P$ indexes all pixels in the input image set $\{\mathbf{I}^{*}_{i}\}_{i=1}^{n}$. The photometric loss is defined on mini-batches of pixels in $P$:
\begin{equation}
    \mathcal{L}_{RGB} = \frac{1}{|P|}\sum_{p\in P^{in}}|\mathbf{I}^*_p-\mathcal{G}(\mathbf{\hat{x}}_p, \mathbf{\hat{n}}_p, \mathbf{f}_p, \mathbf{v}_p)|, 
    \label{eqn:L_rgb_noMap}
\end{equation}
where $\mathbf{f}_p$, $\mathbf{\hat{x}}_p$, $\mathbf{\hat{n}}_p$ are defined in Eqn.~\ref{eqn:g}. $P^{in}\subset P$ represents the subset of pixels $P$ where intersection has been found and $\mathbf{M}_p=1$. $|\cdot|$ denotes the $L_1$ loss. 

\Paragraph{Silhouette Loss} 
We define the silhouette loss as 
\begin{equation}
    \mathcal{L}_{Mask} = \frac{1}{|P|}\sum_{p\in P^{out}}CE(\mathbf{M}_p, \mathbf{\hat{M}}_p),
\end{equation}
where $\mathbf{\hat{M}}$ is the masked rendering. $P^{out}=P - P^{in}$ represents the indices in the mini-batch for which there is no ray-geometry intersection or $\mathbf{M}_p=0$. $CE(\cdot,\cdot)$ denotes the cross-entropy loss. Conventionally, given the ray $\mathbf{r}_p=\{\mathbf{c}_0+t\mathbf{v}_p|t\geqslant 0\}$ of pixel $p$, $\mathbf{\hat{M}}_p$ is defined as:
\begin{equation}
    \mathbf{\hat{M}}_p = \left\{\begin{matrix}
 1& & \mathbf{r}_p\cap \mathcal{S} \\ 
 0& & \textup{otherwise}.
\end{matrix}\right.
\end{equation}
To make this differentiable, 
we follow~\cite{yariv2020multiview} and compute
\begin{equation}
\mathbf{\hat{M}}_p = \textup{sigmoid}(-\beta \, \underset{t\geq0}{\min}\mathcal{F}^{(s)}(\mathbf{c}_0+t\mathbf{v}_p)).
\end{equation}
When $\beta\rightarrow\!\infty $, $\mathcal{F}^{(s)}\!<\!0$ means inside the surface and $\mathcal{F}^{(s)}\!>\!0$ outside.

\Paragraph{Eikonal Regularization}
A special property of signed distance functions is their differentiability with a gradient of unit norm, satisfying the Eikonal equation $||\nabla\mathcal{F}||_2=1$~\cite{osher2004level,gropp2020implicit}. We thus encourage our implicit geometry representation to satisfy the Eikonal property:
\begin{equation}
    \mathcal{L}_{Eik} = \sum_{\mathbf{\tilde{x}}} \begin{Vmatrix} ||\nabla_{\mathbf{\tilde{x}}}\mathcal{F}^{(s)}(\mathbf{\tilde{x}})||_2-1 \end{Vmatrix}_2^2,
\end{equation}
where $\mathbf{\tilde{x}}$ is uniformly sampled at the $3$D region of interest. 

\subsection{Uncertainty Prediction Module}
The pseudo images are inherently aleatoric, {\it i.e.}, there might be areas with either notable artifacts in one image or with inconsistent shape/texture across images. 
To adaptively treat these problematic areas in learning, we propose to use Bayesian learning~\cite{kendall2017uncertainties,xu2021digging} to model this aleatoric uncertainty. 
Specifically, we exploit the feature space of the image encoder $\mathcal{E}$ and train a shallow decoder to estimate an uncertainty map $\mathbf{U}$, which has the same size as pseudo images. 
Formally, we model the observed color $\mathbf{I}^*_p$  at pixel $p$ with a likelihood function $\mathcal{P}(\mathbf{I}^*_p)$ that follows the Laplacian distribution with ray-dependent variance $\sigma_p$:
\begin{equation}
    \mathcal{P}(\mathbf{I}^*_p) = \frac{1}{2\sigma_p} \textup{exp}\left (-\frac{|\mathbf{I}^*_p-\mathcal{G}(\mathbf{\hat{x}}_p, \mathbf{\hat{n}}_p, \mathbf{f}_p, \mathbf{v}_p)|}{\sigma_p}\right ),
\end{equation}
where $\sigma_p$ denotes the uncertainty. Since $L_1$ distance is less sensitive to outliers, which is more suitable for optimizing the rendered appearance against the pseudo RGB values. Thus, we adopt Laplacian likelihood to model the inconsistent uncertainty distribution.
To find the parameters best explaining the model, we maximize the likelihood function, {\it i.e.,}~minimizing the negative log-likelihood:
\begin{equation}
    -\!\textup{log}(\mathcal{P}(\mathbf{I}^*_p)) \!=\! \frac{|\mathbf{I}^*_p\!-\!\mathcal{G}(\mathbf{\hat{x}}_p, \mathbf{\hat{n}}_p, \mathbf{f}_p, \mathbf{v}_p)|}{\sigma_p}\!+\textup{log}\sigma_p\!+\textup{log}2.
\end{equation}

Therefore, we update the photometric loss as:
\begin{equation}
    \mathcal{L}_{RGB} \!= \frac{1}{|P|}  \!\sum_{p\in P_{in}}\!\left( e^{-\mathbf{U}_p}|\mathbf{I}^*_p\!-\mathcal{G}(\mathbf{\hat{x}}_p, \mathbf{\hat{n}}_p, \mathbf{f}_p, \mathbf{v}_p) | + \mathbf{U}_p \right ).
     \label{eqn:L_rgb_Map}
\end{equation}
We train this loss on mini-batches of pixels in $P$. Here $P$ indexes all pixels in the input multi-view image set $\{\mathbf{I}_{i=1}^{*}\}_{i=1}^{n}$, which contribute to the same object's depth, texture and uncertainty learning. During training, for the surface point $\hat{\mathbf{x}_{p}}$, the first term $e^{-\mathbf{U}_p}$ can be seen as a weighted distance which assigns larger weights to less uncertain pixels. The second term $\mathbf{U}_{p}$ is a penalty term. 
$\mathcal{L}_{RGB}$ encourages the neural implicit representation to bring together the texture information of all cross-view corresponding pixels in all images. Consequently, the pixel-wise uncertainty value is able to mine the multi-view inconsistencies among those pixels which contribute to the same $3$D surface point.
In practice, we train a $2$-layer convolutional network to predict the log variance $\mathbf{U}_p:=\textup{log}\sigma_p$ (please refer to \textbf{Supp} for more details).

\subsection{Implementation Details}
The encoder $\mathcal{E}$ is implemented as a ResNet-$18$~\cite{he2016deep} followed by fully-connected layers. 
Both $\mathcal{F}$ and $\mathcal{G}$ have $4$ fully-connected layers. For the main experiment, we set $N{=}2{,}000$, $n{=}40$, $k{=}5$, $W{=}H{=}256$, $d_{f}{=}256$, $\beta=50$, $\lambda_{mask}{=}0.01$, $\lambda_{eik}{=}0.1$.
We implement our model in Pytorch and use the Adam optimizer with a learning rate of $1e{-}4$ for both network and camera pose parameters.

%% file: sec_4_exp.tex
\section{Experimental Results}\label{sec:exp}

\begin{figure}[t]
\newcommand{\tabincell}[2]{\begin{tabular}{@{}#1@{}}#2\end{tabular}}
\resizebox{1\linewidth}{!}{
\begin{tabular}{@{\hspace{-0.01cm}} c @{\hspace{-0.01cm}} c c @{\hspace{-0.01cm}} c @{\hspace{-0.01cm}} c @{\hspace{-0.01cm}} }
   \rotatebox[origin=c]{90}{\scriptsize Image} &  \raisebox{-.5\height}{\includegraphics[scale=0.25]{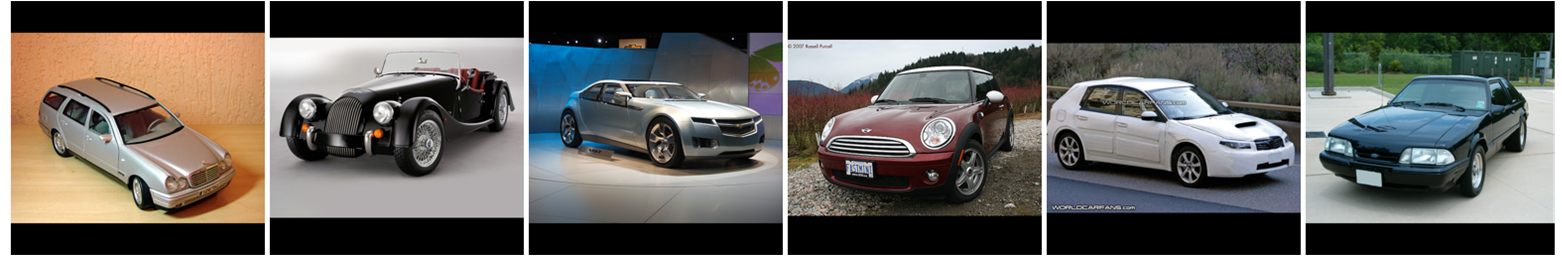}} 
   & &\rotatebox[origin=c]{90}{\scriptsize \tabincell{c}{Human \\ label}}  & \raisebox{-.5\height}{\includegraphics[scale=0.25]{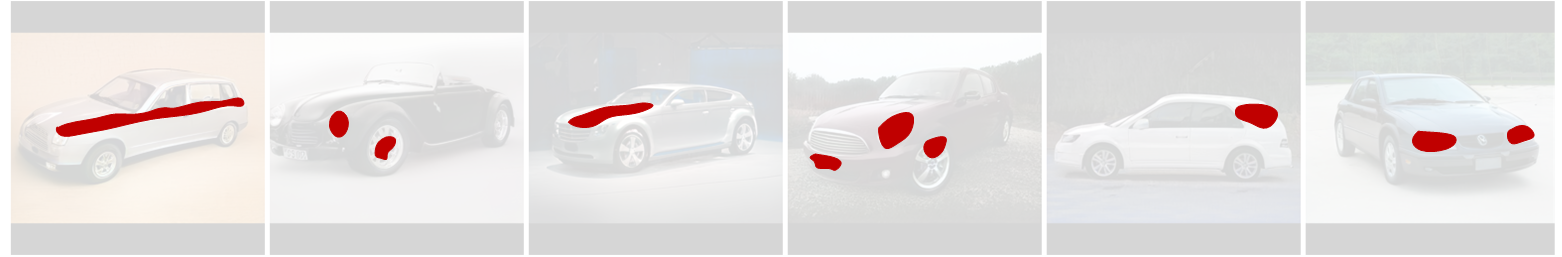}} \\
\rotatebox[origin=c]{90}{\scriptsize \tabincell{c}{Pseudo \\ image} }  &
\raisebox{-.5\height}{\includegraphics[scale=0.25]{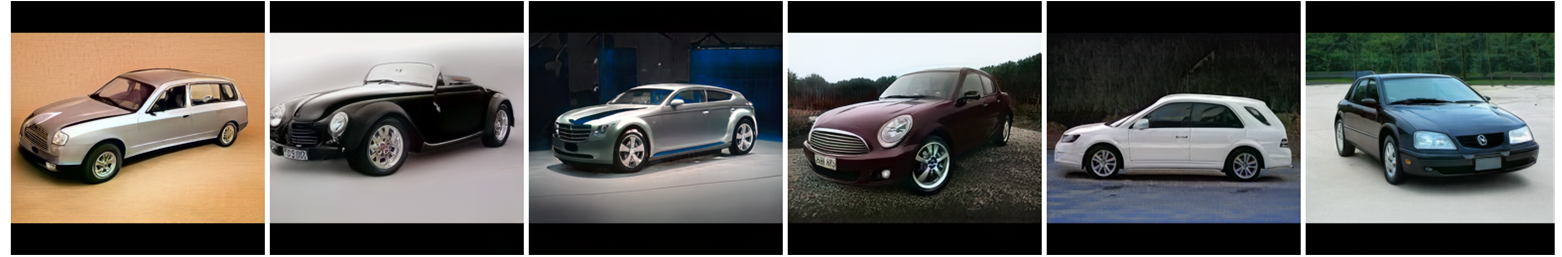}} 
& &
\rotatebox[origin=c]{90}{\scriptsize \tabincell{c}{Our \\ prediction}}  & \raisebox{-.5\height}{\includegraphics[scale=0.25]{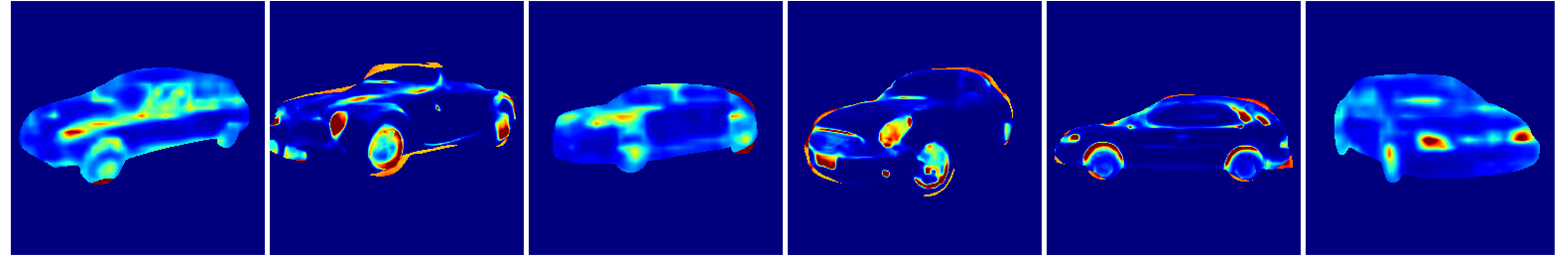}} \\
\end{tabular}
}
\caption{ The uncertainty maps produced by our proposed method and humans.\vspace{-3mm}} 
\label{fig:uncertainty}
\end{figure}

We evaluate our approach on six category-specific StyleGAN models, including rigid objects such as airplanes, cars, motorbikes, as well as non-rigid objects such as birds, horses, and potted plants. 
We use the official car and horse models from StyleGAN2 repo~\footnote{\url{https://github.com/NVlabs/stylegan2}}, trained on the LSUN dataset~\cite{yu2015lsun}. 
For other $4$ categories, we train the StyleGAN models with StyleGAN2-ADA-Pytorch library~\footnote{\url{https://github.com/NVlabs/stylegan2-ada-pytorch}} on LSUN airplanes, birds, motorbikes, and potted plants with $200K$ images respectively.

\subsection{Human Study vs Our Method on the Uncertainty Prediction}
Although GAN interpretation methods have shown that manipulating the latent code of StyleGAN produces multi-view images of the same object~\cite{shen2021closed,harkonen2020ganspace}, no studies have quantitatively evaluated the unreliable/inconsistent object shape or texture in/across the multi-view pseudo images. 
Thanks to our multi-view-stereo-like neural implicit network, 
our uncertainty map can serve as a means to detect the unreliable/inconsistent areas in the GAN-generated multi-view images.
On the other hand, volunteers were asked to label the potentially problematic areas in the pseudo images. As shown in Fig.~\ref{fig:uncertainty}, given an image, the multi-view generator is able to generate pseudo images with varying viewpoints. We believe humans are able to reason unreliable/inconsistent regions in/across the pseudo images. Specifically, this is accomplished using a random set of $100$ images from the PASCAL$3$D+ car category. 
For each image, we generate a pseudo image with a different viewpoint (Sec.~\ref{sec:multi-view}).
Then, the volunteers manually labels polygon-based regions of interest (uncertainty region) on the pseudo images, using the Matlab Image Labeler app. As can be observed in Fig.~\ref{fig:uncertainty}, human labels mainly focus on the global object shape inconsistency across views, which might not have the granularity to evaluate the pixel-level inconsistency.
Nevertheless, we quantify the detection ability of our uncertainty maps by using the human labels as the ground-truth. We achieve $34.6\%$ Intersection over Union (IoU), which shows our capability in detecting the inconsistent object shape in GAN-generated multi-view pseudo images. To our knowledge, this is the first method that tries to investigate the unreliable supervision across the GAN-generated pseudo multi-view images in $3$D object modeling.

\subsection{Quantitative $3$D Reconstruction Evaluation}

We quantitatively evaluate on the PASCAL$3$D+ dataset~\cite{xiang2014beyond}, a $3$D reconstruction benchmark of real-world images with (approximate) CAD model annotations. Similar to prior work~\cite{tulsiani2017multi,lin2020sdf}, we use annotations of airplane and car categories on the test set for evaluation.

\Paragraph{Evaluation Metrics.}
We adopt standard $3$D reconstruction metrics: IoU and Chamfer-$L_1$ Distance (CD). 
Following~\cite{tulsiani2017multi}, we compute $3$D IoU between ground truth and prediction with the resolution of $32^3$.
Following~\cite{lin2020sdf}, we uniformly sample $3$D points from the ground truth and prediction to compute CD.

\Paragraph{Baselines} We compare against SoTA unsupervised single-view $3$D reconstruction baselines: CSDM~\cite{tulsiani2016learning}, DRC~\cite{tulsiani2017multi}, CMR~\cite{kanazawa2018learning}, 
U-CMR~\cite{goel2020shape} and SDF-SRN~\cite{lin2020sdf}. 
As detailed in Tabs.~\ref{tab:3D_modeling_review} and~\ref{tab:pascal_results}, some baselines require additional unsupervised constraints, \emph{e.g.}, DRC (implicit) and SDF-SRN (implicit) both require ground-truth camera pose for each training sample. 
The mesh-based methods such as CSDM, CMR, and U-CMR need expert object-specific templates as additional constraints. 
Here, we do not compare with ShSMesh~\cite{ye2021shelf} as it neither quantitatively evaluates on PASCAL$3$D+, nor trains on real-world car/airplane images.
Also, we do not quantitatively compare with StyleGANRender~\cite{zhang2020image} as the code or trained model is not publicly available, and StyleGANRender does not report full $3$D shape reconstruction errors as our baselines did.

\begin{table}[t]
     \caption{ (a) Quantitative $3$D reconstruction results on PASCAL$3$D+. 
During training, CSDM, DRC and SDF-SRN require ground-truth camera pose per training sample, CMR uses $2$D keypoints and object-specific templates as additional constraints, and U-CMR only relies on object-specific templates. [Keys: Requ. or Cons.=requirement or constraint in training, T=category-specific templates, C=poses per training sample, $\textup{C}^*$=poses for reference images,  K=$2$D keypoints, S=semantic information, GANs=pre-trained GAN models]. All CD values are scaled by $10$ following~\cite{lin2020sdf}. (b) Ablation of uncertainty prediction and the number of pseudo images, using the PASCAL$3$D+ car category.\vspace{-2mm}}
    \begin{subtable}{0.48\textwidth}
        \centering
        
        \caption{}
        \vspace{-2mm}
 \newcommand{\tabincell}[2]{\begin{tabular}{@{}#1@{}}#2\end{tabular}} 
 \resizebox{1\linewidth}{!}{
\begin{tabular}{l| c| c  | c | c | c}
\hline
\multirow{ 2}{*}{Category}   & \multirow{ 2}{*}{\tabincell{c}{Requ.  \\ or Cons.} }   &\multicolumn{2}{c|}{Airplane}  & \multicolumn{2}{c}{Car}    \\  
\cline{3-4}
\cline{5-6}
& &CD ($\downarrow$)&IoU ($\uparrow$)& CD ($\downarrow$)&IoU ($\uparrow$)\\
\hline 
\hline 
CSDM~\cite{tulsiani2016learning}  &  T, C   &  $-$ & $0.400$   &  $-$  & $0.600$ \\  
DRC~\cite{tulsiani2017multi}      &  C   &  $-$ & $0.420$   &  $-$  & $0.670$ \\
CMR~\cite{kanazawa2018learning}   &  T, K &  $0.625$ & $-$      &  $0.474$  & $0.640$ \\
U-CMR~\cite{goel2020shape}        &  T   &  $-$ & $-$      &  $-$  & $0.646$ \\
UMR~\cite{li2020self}             &   S  &  $-$ & $-$      &  $-$  & $0.620$  \\

SDF-SRN~\cite{lin2020sdf}       &  C    &  $0.303$ & $0.405$ &  $0.233$ & $0.653$ \\
SDF-SRN*~\cite{lin2020sdf}       &  C    &  $0.297$ & $0.412$ &  $0.230$ & $0.661$\\
\hline
\textbf{Proposed}   &   GANs, $\textup{C}^*$   & $\textbf{0.286}$ & $\textbf{0.473}$ &  $\textbf{0.195}$ & $\textbf{0.702}$ \\
\hline 
\end{tabular}
}
       
       \label{tab:pascal_results} 
    \end{subtable}
    \begin{subtable}{0.48\textwidth}
        \centering
        \caption{}
         \newcommand{\tabincell}[2]{\begin{tabular}{@{}#1@{}}#2\end{tabular}} 
\resizebox{1\linewidth}{!}{
\begin{tabular}{l| c| c | c | c | c}
\hline
 &  \multirow{ 2}{*}{\tabincell{c}{\textbf{Proposed}  w/o\\ Uncertainty ($n{=}40$)} } 
 & \multicolumn{4}{c}{\textbf{Proposed} ($n=$)} \\
 \cline{3-6}
 & & $10$ & $20$ & $40$ & $50$ \\
 
 \hline
 \hline
 CD ($\downarrow$) & $0.208$ & $0.336$ & $0.243$ & $0.195$ & $\textbf{0.191}$\\ \hline
 IoU ($\uparrow$)& $0.681$ & $0.612$ & $0.643$ & $0.702$ & $\textbf{0.714}$\\
 
\hline 
\end{tabular}
}
         \vspace{-4mm}
        \label{tab:ablation}
     \end{subtable}
\end{table}

\begin{figure}[t]
\centering
\newcommand{\tabincell}[2]{\begin{tabular}{@{}#1@{}}#2\end{tabular}}
\resizebox{0.9\linewidth}{!}{
\begin{tabular}{@{\hspace{-0.01cm}} c @{\hspace{-0.01cm}} c @{\hspace{-0.01cm}}}
   \rotatebox[origin=c]{90}{\scalebox{.5}{ Input}} &  \raisebox{-.5\height}{\includegraphics[scale=0.25]{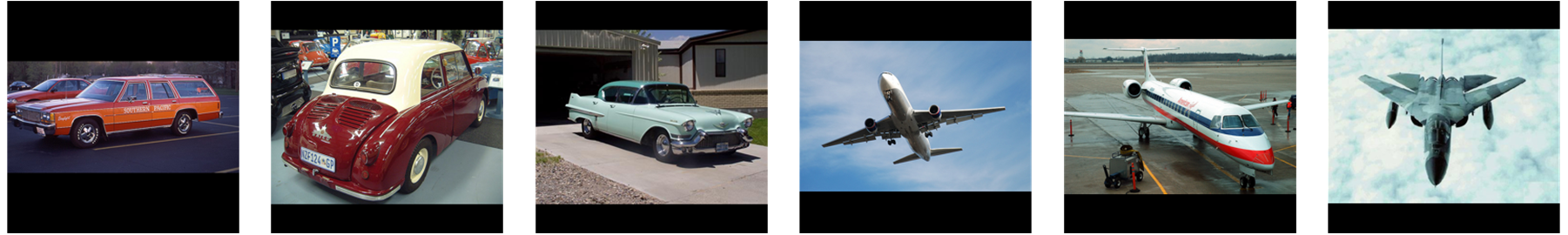}} \\
\rotatebox[origin=c]{90}{ \scalebox{.4}{  SDF-SRN} }  &
\raisebox{-.5\height}{\includegraphics[scale=0.25]{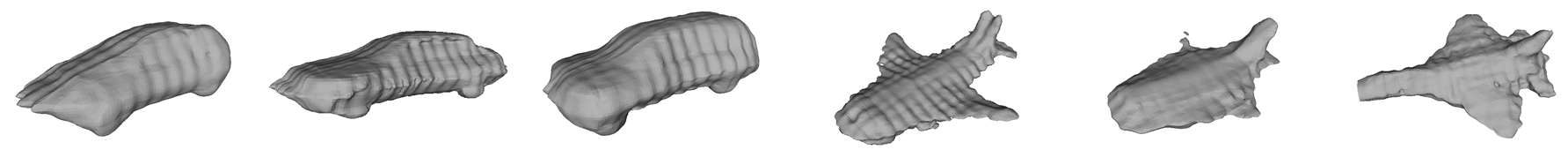}} \\
\rotatebox[origin=c]{90}{ \scalebox{.4}{ \textbf{Proposed}}}  & \raisebox{-.5\height}{\includegraphics[scale=0.25]{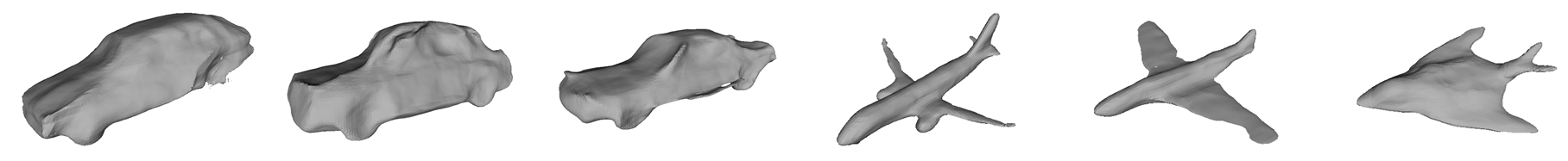}} \\
\rotatebox[origin=c]{90}{ \scalebox{.4}{ \textbf{GT}}}  & \raisebox{-.5\height}{\includegraphics[scale=0.25]{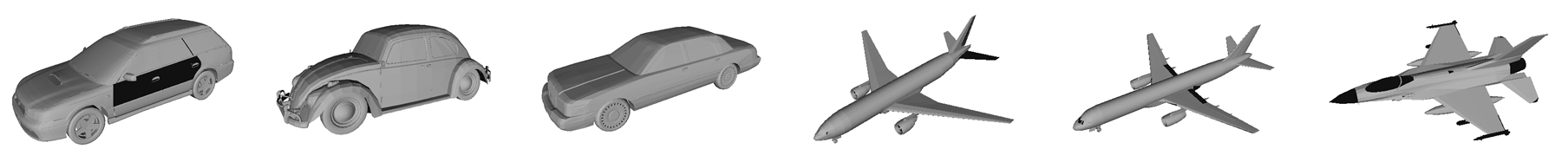}} \\ \hline
& \raisebox{-.5\height}{\includegraphics[scale=0.25]{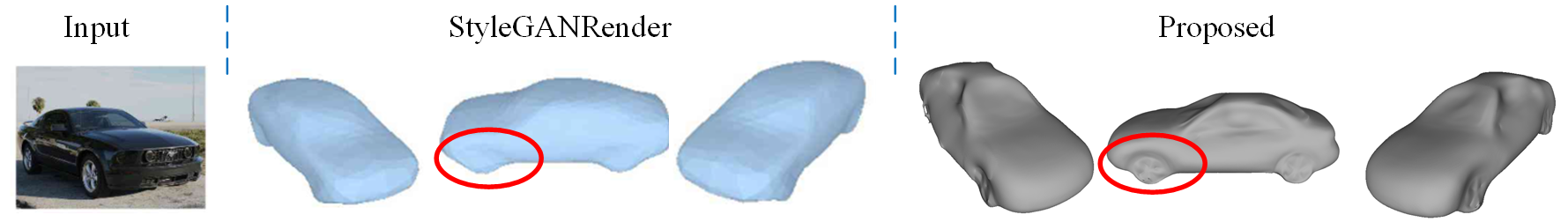}} \\
\end{tabular}
}
\caption{Qualitative comparisons with SDF-SRN~\cite{lin2020sdf} (SOTA baseline) and  StyleGANRender~\cite{zhang2020image} on PASCAL$3$D+ car or airplane categories. Our approach recovers significantly more accurate $3$D shapes and topologies from the images.}
\vspace{-2mm}
\label{fig:pascal_results}
\end{figure}

\Paragraph{Results}
We present the comparisons of our approach in Tab.~\ref{tab:pascal_results} and visualize sample predictions in Fig.~\ref{fig:pascal_results}. 
It can be observed that the \textbf{Proposed} model is significantly better than baselines in both CD ($10.1\%$ relative over SDF-SRN) and IoU ($7.8\%$ relative over DRC). It is worth noting that our models trained with Proposed setting only require ground-truth pose annotations for the reference images.
It is more practical for real-world scenarios than DRC and SDF-SRN, which require ground-truth camera pose per training sample. 
Further, we re-train SDF-SRN with our GAN-generated training data and report the results in Tab.~\ref{tab:pascal_results} (SDF-SRN*). It can be observed, despite the minor improvement over the original SDF-SRN due to our pseudo images, the new model still performs worse than ours.
%
%
Fig.~\ref{fig:pascal_results} shows visual comparisons to SDF-SRN~\cite{lin2020sdf} and StyleGANRender~\cite{zhang2020image} results. 
As can be observed, our approach suffers slightly from shape ambiguity, \emph{e.g.}, windows of cars tend to be concave. Nonetheless,  our predictions more closely resemble the ground truth.


\subsection{Ablation Study} 
All ablations use the models trained on the car category.

\Paragraph{Effect on Uncertainty Prediction} 
We evaluate single-view $3$D reconstruction on a model trained without uncertainty prediction, {\it i.e.}~using Eqn.~\ref{eqn:L_rgb_noMap} instead of Eqn.~\ref{eqn:L_rgb_Map}. 
As shown in Tab.~\ref{tab:ablation}, our uncertainty prediction module can remedy the negative impact of uncertain texture in GAN-generated multi-view images, leading to improved $3$D reconstruction (CD: $0.208\rightarrow0.195$).

\Paragraph{Effect on $n$}
The key assumption, as well as motivation of our work, is that GAN-generated multi-view images can be leveraged to learn a multi-view stereo system. 
To validate the impact of the amount of pseudo images, we train models with different numbers of image viewpoints, $n=10,20,40,50$. Tab.~\ref{tab:ablation} shows that the model trained with $n=40$ images significantly outperforms the ones with $n=10,20$, and saturates when $n=40\xrightarrow[]{}50$. 
Considering the tradeoff between reconstruction accuracy and training cost, we use $n=40$ for all experiments.

\Paragraph{Effect on Multi-view Generation}
Following the same setting, we re-train a model with pseudo imaged produced by \emph{\textbf{Baseline}} method (Sec.~\ref{sec:multi-view}). Quantitatively, such a model only obtain the IoU of $0.679$, much worse than ours ($0.702$). The comparisons show the superior quality of our multi-view generation method.    




\subsection{Qualitative Evaluation}

\begin{figure}[t]
\newcommand{\tabincell}[2]{\begin{tabular}{@{}#1@{}}#2\end{tabular}}
\resizebox{1\linewidth}{!}{
\begin{tabular}{@{\hspace{-0.01cm}} c @{\hspace{-0.01cm}} c  @{\hspace{-0.01cm}} c  c  @{\hspace{-0.01cm}} c  @{\hspace{-0.01cm}}}
   \rotatebox[origin=c]{90}{\tiny Input} &  \raisebox{-.5\height}{\includegraphics[scale=0.25]{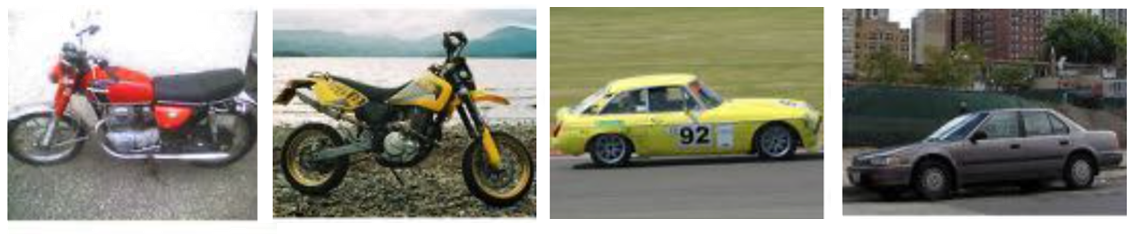}} & & \rotatebox[origin=c]{90}{\tiny Input} & \raisebox{-.5\height}{\includegraphics[scale=0.25]{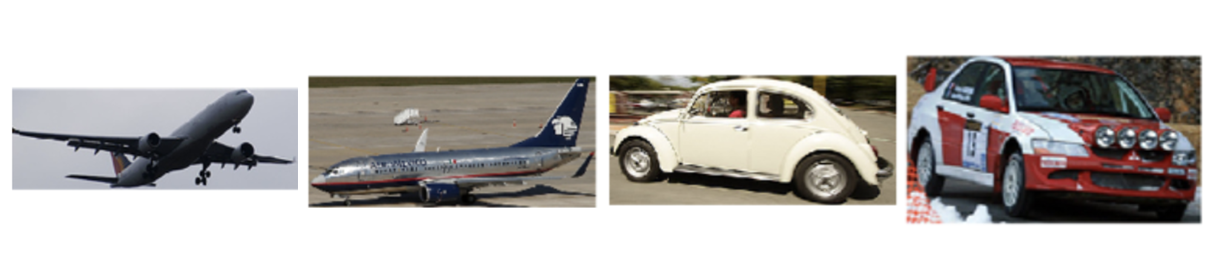}}  \\
\rotatebox[origin=c]{90}{\tiny U-CMR~\cite{goel2020shape} }  &
\raisebox{-.5\height}{\includegraphics[scale=0.25]{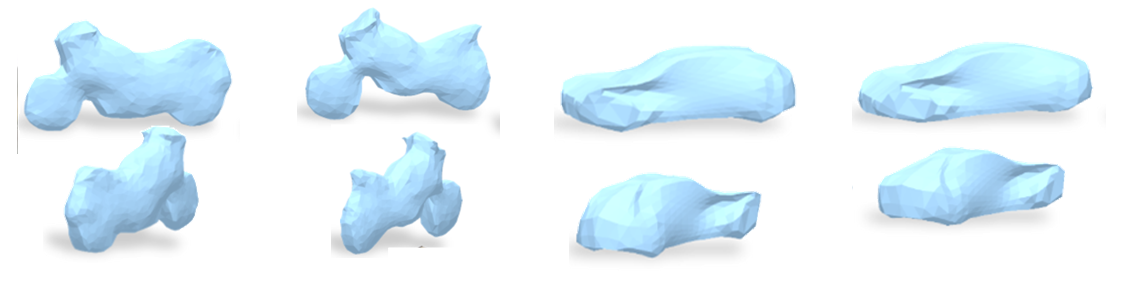}} & &  \rotatebox[origin=c]{90}{\tiny DRC~\cite{tulsiani2017multi} } & \raisebox{-.5\height}{\includegraphics[scale=0.25]{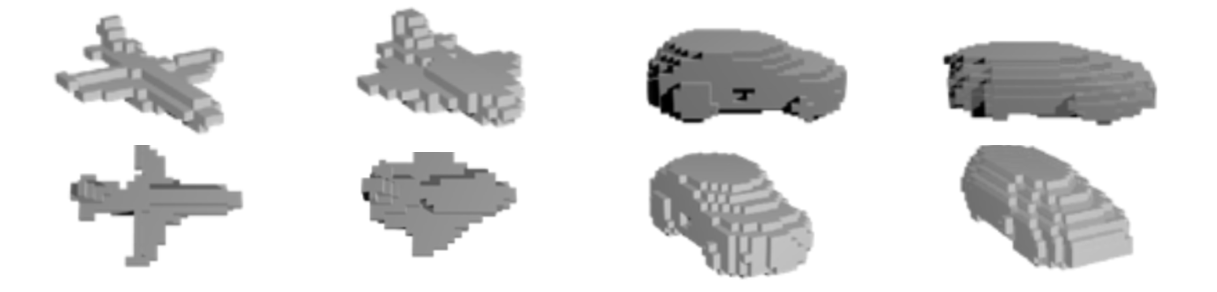}} \\
\rotatebox[origin=c]{90}{\tiny Proposed}  & \raisebox{-.5\height}{\includegraphics[scale=0.25]{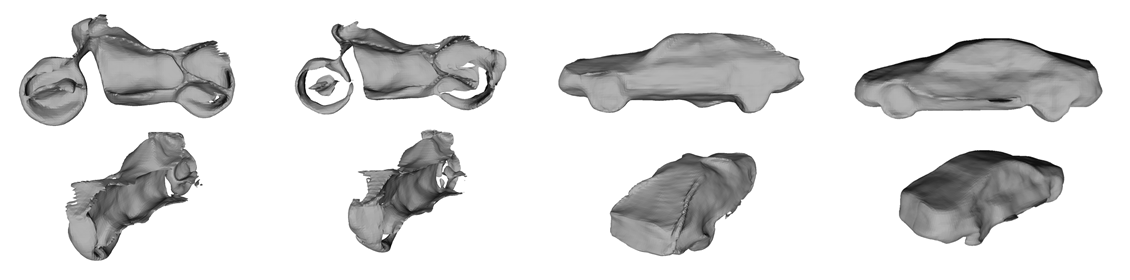}} &  & \rotatebox[origin=c]{90}{\tiny Proposed}  &  \raisebox{-.5\height}{\includegraphics[scale=0.25]{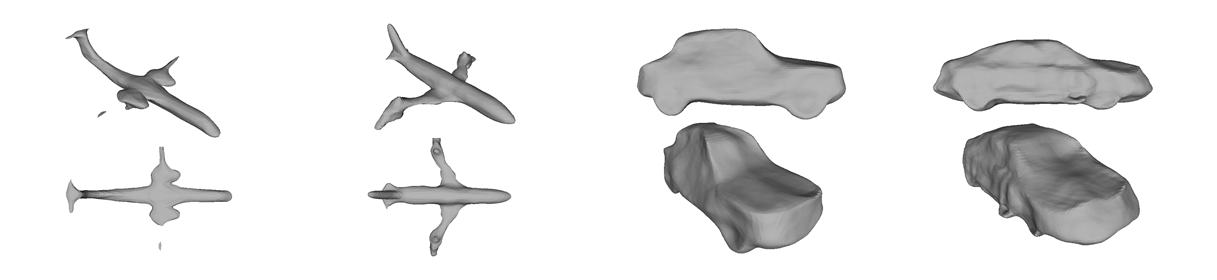}}\\
\end{tabular}
}
\caption{ Additional qualitative comparisons with U-CMR~\cite{goel2020shape} and DRC~\cite{tulsiani2017multi} on cars and motorcycles or airplanes.}
\label{fig:visual_comparison}
\end{figure}

\Paragraph{Comparison with U-CMR~\cite{goel2020shape} and DRC~\cite{tulsiani2017multi}}
We show qualitative comparisons with U-CMR and DRC on the PASCAL$3$D+ cars, motorbikes or airplanes, in Fig.~\ref{fig:visual_comparison}. 
Our approach achieves more faithful reconstructions than U-CMR. Note that U-CMR requires expert templates while our approach does not.

\begin{figure}[t]
\centering
\newcommand{\tabincell}[2]{\begin{tabular}{@{}#1@{}}#2\end{tabular}}
\resizebox{1\linewidth}{!}{
\begin{tabular}{@{\hspace{-0.01cm}} c @{\hspace{-0.01cm}} c @{\hspace{-0.01cm}}}
   \rotatebox[origin=c]{90}{\scalebox{.5}{\large Input}} &  \raisebox{-.5\height}{\includegraphics[scale=0.25]{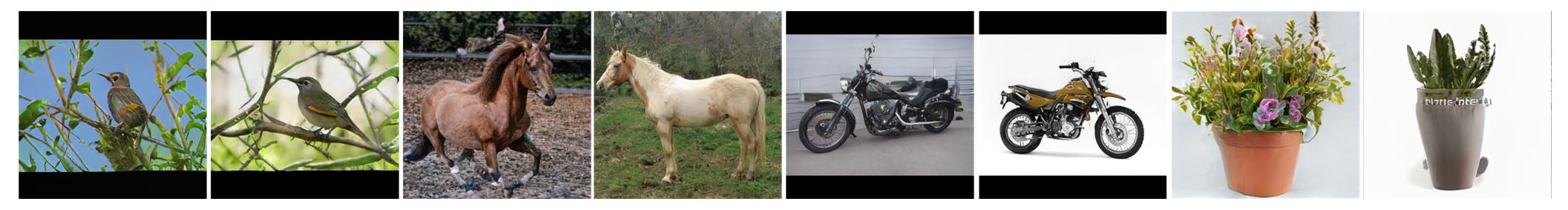}} \\
\rotatebox[origin=c]{90}{ \scalebox{.4}{\large  Proposed} }  &
\raisebox{-.5\height}{\includegraphics[scale=0.25]{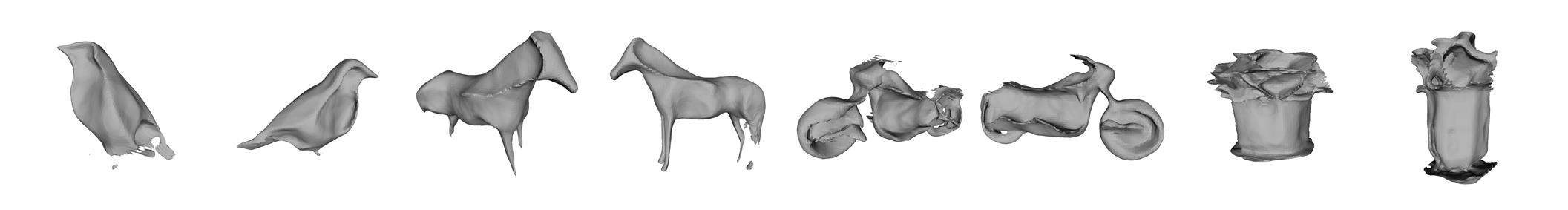}} \\ 
\rotatebox[origin=c]{90}{ \scalebox{.4}{\large Input}}  & \raisebox{-.5\height}{\includegraphics[scale=0.25]{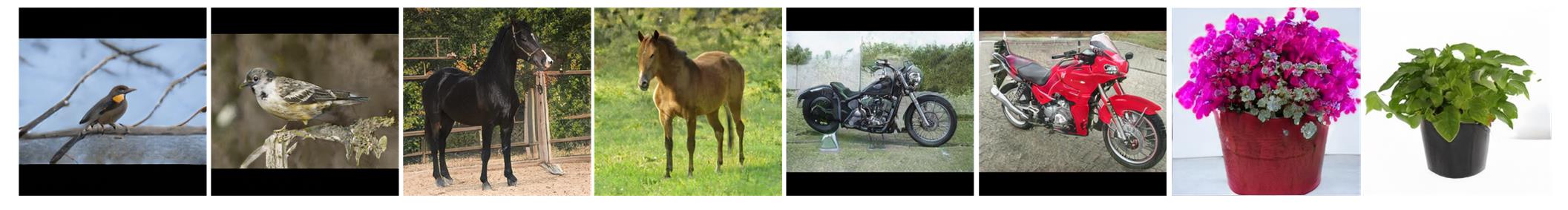}} \\
\rotatebox[origin=c]{90}{ \scalebox{.4}{\large Proposed}}  & \raisebox{-.5\height}{\includegraphics[scale=0.25]{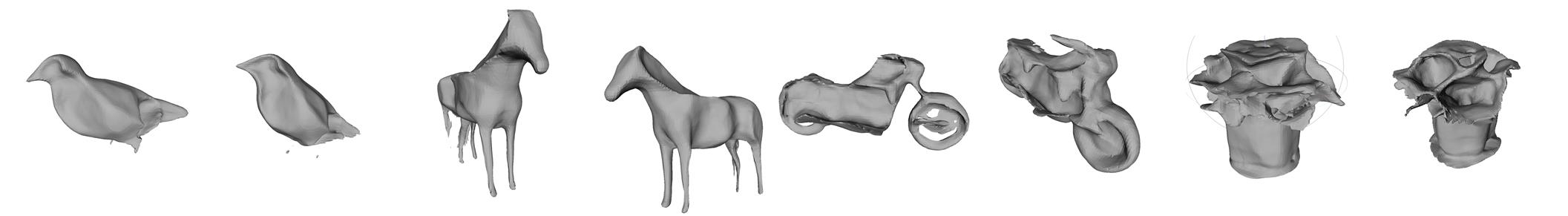}} \\ 
\end{tabular}
} 
\caption{Qualitative results of birds, horses, motorbikes, and potted plants.}
\label{fig:add_results}
\end{figure}

\Paragraph{Results on More Categories}
While our quantitative evaluation is on the PASCAL$3$D+ airplane and car, we provide more qualitative results for the birds, horses, motorbikes and potted plants in Fig.~\ref{fig:add_results}. As can be seen, our approach can effectively capture the thin structure present in $3$D shapes from single-view images, {\it e.g.}~horses' legs and motorbikes' hand clutch.
All testing images are from the LSUN dataset, which never appear in the training set of our GAN models.

%% file: sec_5_conclusion.tex
\section{Conclusions}\label{sec:con}
To leverage pre-trained GAN models for $3$D vision tasks, we propose an image-conditioned neural implicit network that can 
learn the shape priors from GAN-generated multi-view images and perform single-view $3$D reconstruction.  
Moreover, we naturally introduce a novel uncertainty prediction module, which can avoid invalid supervisions for better single-view $3$D reconstruction. Experimentally, our approach significantly outperforms the SoTA unsupervised single-view $3$D reconstruction methods, while requiring less supervision during training. 
We believe this work opens up a path for improving the ability to semantically control GAN generation and facilitates $2$D GAN priors for $3$D vision tasks.

\Paragraph{Limitations} For some categories (\emph{e.g.}, chair), StyleGAN is unable to converge to satisfying results, partially due to chairs' large topology variations. 
We believe the rapid development of GANs will extend to these challenging categories, and thus our method can leverage them for $3$D reconstruction of more categories.
Also, similar to most prior works, our model is category-specific.
One future direction is to develop a single model for multiple categories based on universal GAN models, \emph{e.g.}, BigGAN~\cite{brock2018large}, improving generalization to unseen categories.